\pdfoutput=1

\documentclass[11pt]{article}
\usepackage{epigraph}

\usepackage{booktabs}
\usepackage{enumitem}
\usepackage{balance}

\usepackage{ACL2023}
\usepackage{graphicx} 

\usepackage{multirow}

\usepackage{times}
\usepackage{mathptmx}
\usepackage{latexsym}
\usepackage{amsmath,amsthm,amssymb, graphicx, multicol, array, mathtools}
\usepackage{xcolor} 
\usepackage{hyphenat}

\usepackage[T1]{fontenc}

\usepackage[utf8]{inputenc}

\usepackage{microtype}

\usepackage{inconsolata}

\newcommand{\ignore}[1]{}

\newcommand{\upperestimate}{46\%}

\newcommand{\xhdr}[1]{\vspace{1.7mm}\noindent{{\bf #1.}}}

\title{Artificial Artificial Artificial Intelligence: Crowd Workers Widely Use Large Language Models for Text Production Tasks}

\author{Veniamin Veselovsky,$^*$
Manoel Horta Ribeiro,$^*$
Robert West\\
    EPFL\\ \texttt{firstname.lastnames@epfl.ch}    }

\begin{document}

\maketitle
\def\thefootnote{*}\footnotetext{Equal contribution.\\ }
\def\thefootnote{\arabic{footnote}}
\begin{abstract}

Large language models (LLMs) are remarkable data annotators. They can be used to generate high-fidelity supervised training data, as well as survey and experimental data.
With the widespread adoption of LLMs, human gold\hyp standard annotations are key to understanding the capabilities of LLMs and the validity of their results.
However, crowdsourcing, an important, inexpensive way to obtain human annotations, may itself be impacted by LLMs, as crowd workers have financial incentives to use LLMs to increase their productivity and income.
To investigate this concern, we conducted a case study on the prevalence of LLM usage by crowd workers.
We reran an abstract summarization task from the literature on Amazon Mechanical Turk and, through a combination of keystroke detection and synthetic text classification, estimate that 33--\upperestimate{} of crowd workers used LLMs when completing the task.
Although generalization to other, less LLM-friendly tasks is unclear, our results call for platforms, researchers, and crowd workers to find new ways to ensure that human data remain human, perhaps using the methodology proposed here as a stepping stone.
Code/data:  \url{https://github.com/epfl-dlab/GPTurk}
\end{abstract}



\section{Introduction}
\epigraph{Normally, a human makes a request to a computer, and the computer does the computation of the task. But \textbf{artificial artificial intelligences} like Mechanical Turk invert all that.}{Jeff Bezos}

\noindent
What do massive computer vision datasets \cite{deng2009imagenet} and large-scale psychological experiments~\cite{pennycook2021shifting} have in common? 
Both rely on crowd work done on platforms such as Amazon Mechanical Turk (MTurk), Prolific, or Upwork.
Such crowdsourcing platforms have become central to researchers and industry practitioners alike, offering ways to create,  annotate, and summarize all sorts of data~\cite{gray2019ghost,schwartz2019untold}, and to run surveys
surveys as well as experiments~\cite{salganik2019bit}.

At the same time, large language models (LLMs), including ChatGPT, GPT-4, PaLM, and Claude, have taken the digital world by storm. Early work indicates that LLMs are remarkable data annotators, outperforming both crowd workers ~\cite{kocon2023chatgpt,gilardi2023chatgpt} and experts~\cite{tornberg2023chatgpt}.
Moreover, they show promise in simulating human behavior, allowing social scientists to conduct \textit{in silico} experiments and surveys, obtaining similar results as would be obtained from real humans \cite{argyle2022out, horton2023large, dillion2023can}.
Yet, human experimental subjects, annotators, and survey-takers remain critical of the validity of results derived from LLMs, as
 they  still perform poorly at various tasks~\cite{ziems2023can} and synthetic data generated by LLMs can still be unfaithful with respect to the real data of interest \cite{veselovsky2023generating}. 

\begin{figure*}
    \centering
    \includegraphics[width=0.75\textwidth]{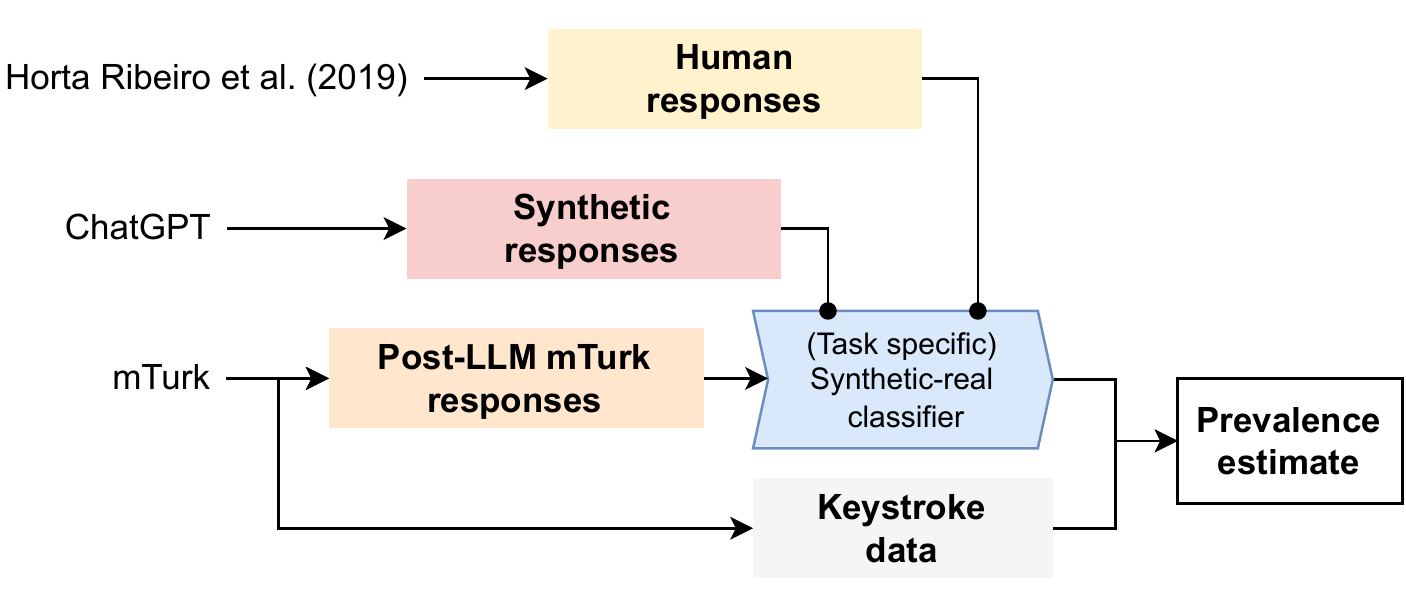}
    \caption{Illustration of our approach for quantifying the prevalence of LLM usage among crowd workers solving a text summarization task. First, we use truly human-written MTurk responses and synthetic LLM-written responses to train a task-specific synthetic-vs.-real classifier. Second, we use this classifier on real MTurk responses (where workers may or may not have relied on LLMs), estimating the prevalence of LLM usage. 
    Additionally (not shown), we confirm the validity of our results in a post-hoc analysis of keystroke data collected alongside MTurk responses.}
    \label{fig:idea}
\end{figure*}

Given this situation, it is tempting to rely on crowdsourcing to validate LLM outputs or to create human gold-standard data for comparison. 
But what if crowd workers themselves are using LLMs, e.g., in order to increase their productivity, and thus their income, on crowdsourcing platforms?
We argue that this would severely diminish the utility of crowdsourced data because the data would no longer be the intended human gold standard, but also because one could prompt LLMs directly (and likely more cheaply) instead of paying crowd workers to do so (likely without disclosing it).

For these reasons, we are curious to what extent crowd workers are already using LLMs as part of their work.
The answer to this question is of paramount importance for all who rely on crowdsourcing, for without knowing who really produced the data---humans or machines---it is hard to assess in what ways one can rely on the data.
As a first step toward answering this question, we here quantify the usage of LLMs by crowd workers through a case study on MTurk, based on a novel methodology for detecting synthetic text.
In particular, we consider part of the text summarization task from \citet{horta2019message}, where crowd workers summarized 16 medical research paper abstracts.
By combining keystroke detection and synthetic\hyp text classification, we estimate that 33--\upperestimate{} of the summaries submitted by crowd workers were produced with the help of LLMs.

We conclude that, although LLMs are still in their infancy, textual data collected via crowdsourcing is already produced to a large extent by machines, rather than by the hired human crowd workers.
Although our study specifically considers a text summarization task, we caution that any text production task whose instructions can be readily passed on to an LLM as a prompt are likely to be similarly affected.
Moreover, LLMs are becoming more popular by the day, and multimodal models, supporting not only text, but also image and video input and output, are on the rise. 
With this, our results should be considered the ``canary in the coal mine'' that should remind platforms, researchers, and crowd workers to find new ways to ensure that human data remain human.


\section{Related work}


\vspace{-1mm}
\xhdr{Research using crowdsourcing}
When Amazon Mechanical Turk was first released, it was colloquially known as ``artificial artificial intelligence'' (see opening quote by Jeff Bezos), since to requesters (users who upload tasks to the platform) it seemed as though an artificial intelligence system was solving their tasks, although, in reality, humans did \cite{schwartz2019untold}. 
From transcription~\cite{marge2010using} to image annotation~\cite{sorokin2008utility}, MTurk spearheaded a paradigm shift in how machine learning datasets were created~\cite{paullada2021data, marge2010using} and how user studies~\cite{sorokin2008utility}, surveys~\cite{buhrmester2011amazon}, and social science experiments~\cite{pennycook2021shifting, salganik2019bit} were conducted.

\vspace{-1mm}
\xhdr{Research about crowdsourcing}
The ubiquity of MTurk and the many crowdsourcing platforms that have since appeared led to a rich body of literature about crowdsourcing.
Previous work has studied how to efficiently use crowdsourcing, with all its limitations, to accomplish a variety of tasks~\cite{draws2021checklist, bragg2013crowdsourcing}, conducted audits on the overall quality of crowdsourced annotations~\cite{chmielewski2020mturk,kennedy2020shape,smith2016multi},
and shed light on the demographics and socioeconomic conditions of workers using crowdsourcing platforms~\cite{burnham2018mturk,mccredie2019turkers,ogletree2021older,ipeirotis2010demographics,gray2019ghost}.

\vspace{-1mm}
\xhdr{LLM-generated data}
As previously mentioned, LLMs can act as effective proxies for human sub-populations~\cite{argyle2022out}, leading to a series of studies using LLMs as ``silicon samples''~\cite{argyle2022out,horton2023large,dillion2023can}. 
Typically, these analyses have been done through a variant of controlled text generation (see \citet{zhang2022survey} for a comprehensive review). Further, an ever-increasing body of work illustrates the good performance of LLMs as proxies for human labeling~\cite{wang2023chatgpt,gilardi2023chatgpt,ziems2023can} and for generating high-quality text (which is, however, often factually inaccurate) that has captured the imagination of the general public and the media~\cite{dale2021gpt}.

\vspace{-1mm}
\xhdr{Detecting LLM-generated data}
Distinguishing LLM- from human-generated text is difficult for both machine learning models and humans alike~\cite{sadasivan2023can, jakesch2023human}.
For example, OpenAI's own LLM-vs.-human classifier recognizes only 26\% of LLM-written texts as such.%
\footnote{\url{https://openai.com/blog/new-ai-classifier-for-indicating-ai-written-text}}
Thus, in the context of the explosive popularity of LLMs such as ChatGPT, there is widespread concern about their usage in areas such as social media~\cite{pan2023risk} (where they could be used to generate spam or misinformation) or higher education~\cite{rudolph2023war} (where they could be used to cheat assignments and exams).
These concerns have led to work on watermarking LLM output~\cite{kirchenbauer2023watermark} by slightly altering the odds of each token during sampling, and to further work on methods for improving the detection of synthetically generated text~\cite{yu2023cheat, verma2023ghostbuster, diwan2021fingerprinting}.

\begin{figure*}[t]
    \centering
    \includegraphics[width=.99\textwidth]{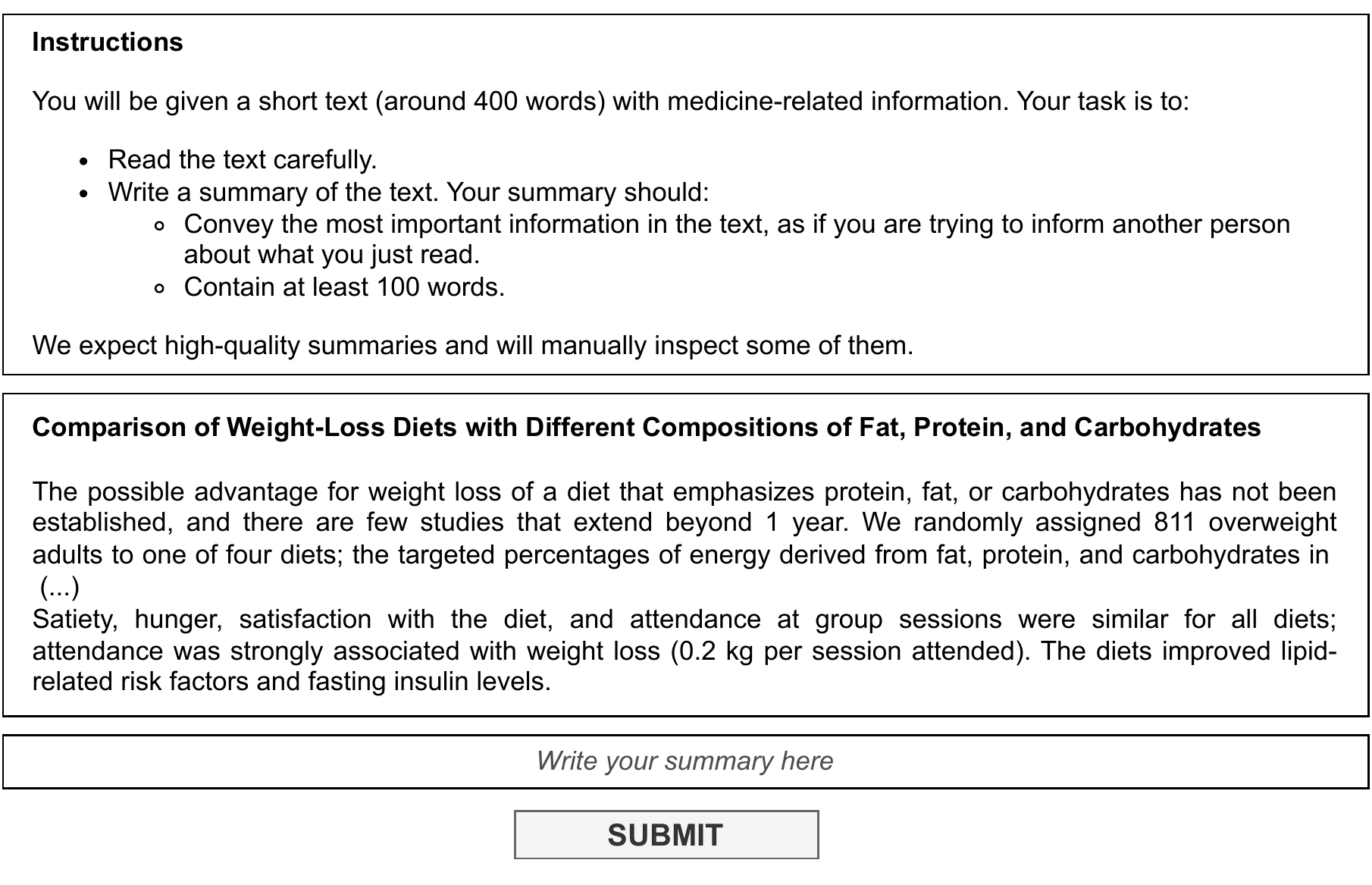}
    \caption{Depiction of the MTurk task studied in this paper, where crowd workers were asked to condense research abstracts from the \textit{New England Journal of Medicine} into summaries about 100 words long.}
    \label{fig:hil}
\end{figure*}

\section{Methods}

We illustrate our overall approach in Figure~\ref{fig:idea} and describe it in detail below.

\subsection{Task of choice: abstract summarization}
We modify a prior MTurk task originally devised by \citet{horta2019message}, whose goal was to study the so-called ``telephone effect,'' whereby information is gradually lost or distorted as a message is passed from human to human in an information cascade.
As part of their experiment, crowd workers were given medical research paper abstracts published in the \textit{New England Journal of Medicine} (NEJM) and were asked to summarize the original abstract (about 2,000 characters) into a much shorter paragraph (about 1,000 characters). The process was then iterated with the summaries, rather than the original abstracts, as the input texts to be summarized further, and so on for multiple rounds.
The original abstracts were about four research topics of  public interest (vaccination, breast cancer, cardiovascular disease, and nutrition), and four papers were selected per topic, for a total of 16 abstracts.
We chose this task for our study for two reasons. First, it is laborious for humans while being easily done with the aid of commercially available LLMs~\cite{luo2023chatgpt}. Second, it is a good example of a task where truly human text is fundamentally required: the very point of \citet{horta2019message} was to study how information is lost when \textit{humans} summarize text, which would not have been possible with synthetically generated, rather than human-generated, data.

In the original study, crowd workers produced eight increasingly short summaries of each original abstract, forming entire information cascades. For our purpose, however, we reduced the task to a single summarization step, where an abstract was condensed into a concise summary of ideally about 100 words (exmaple in Figure~\ref{fig:hil}).
Workers were paid 1~USD per summary, which we estimate conservatively would take around 4 minutes to conclude, for a pay rate of 15~USD/hour.

We obtained 48 summaries written by 44 distinct workers. For two of the abstracts, two summaries were duplicates, which we de-duplicated, leaving us with 46 summaries. Summaries were written around 1~June 2023. Besides the summaries, we also used Javascript to extract all keystrokes made by workers while performing the task, including copy and paste actions (with the Ctrl+C and Ctrl+V keyboard shortcuts or without the keyboard, e.g., using the menu appearing after a right-click).

\textit{\textbf{Note:} This is work in progress. We continue to run the task and will update this pre-print accordingly.}

\subsection{Detecting synthetic text}
\label{sec:det}
To estimate the prevalence of LLM usage in the summarization task described, we need to detect whether the answers provided by crowd workers were synthetically generated.
 Out-of-the-box solutions such as GPTZero,
 OpenAI's AI Detector,
 or Writer,%
 \footnote{%
 \url{https://gptzero.me/},
 \url{https://platform.openai.com/ai-text-classifier},
 \url{https://writer.com}
 }
 work well on simple texts, but in our context, these methods fail to perform effectively; e.g., out of 10 summaries that we had synthesized via ChatGPT, GPTZero detected only six as synthetic.

 Consequently, we rely on a more bespoke solution for detecting LLM\hyp generated summaries by fine-tuning our own model to detect the usage of ChatGPT, which at the time of writing is the most commonly used LLM. Although we only consider a few summaries in this paper, this approach is also more tractable for larger-scale future datasets where API calls may be expensive and slow. 
 
\xhdr{Model architecture} We used an \texttt{e5-base} pre-trained model as our main architecture~\cite{wang2022text}. This model was pre-trained using a contrastive loss and achieves strong performance in a fine-tuned classification setting. 
During fine-tuning, we used a learning rate of $2\times 10^{-5}$, a batch size of 32, a max token length of 256, and trained for five epochs, saving for later use the best-performing model on the validation set.  

\xhdr{Data}
To train the classifier, we employed three datasets, all originating or derived from the MTurk task in question (see Figure~\ref{fig:idea}).
The real instances of human text included the 16 original abstracts and the real human responses (128 high-quality summarizations) from \citeauthor{horta2019message}'s version of the MTurk task. We generated the synthetic instances by prompting ChatGPT,%
\footnote{\url{https://openai.com/blog/chatgpt}}
using the MTurk task instruction as the prompt, which we suspect many crowd workers also did, since they often copied this instruction, as detected via keystroke logging.
For each abstract, we generated 10 different summaries for each of two temperature values (0.7 and 1), obtaining 320 synthetic samples.


\xhdr{Training}
We trained the model in two train/test setups: an \emph{abstract-level} split and  a \emph{summary-level} split. In the abstract-level split, we divide the abstracts into two disjoint sets: 12 abstracts for training and validation, and four for testing, totaling 370 training points. 
The abstract-level split serves as a basis for how well the model is able to extract generalizable artifacts present in the synthetic text that can be exploited for detecting synthetic summaries of abstracts not seen during training. 
In the summary-level split, we randomly split both the real and synthetic summary datasets, utilizing 75\% of the summaries for training, 10\% for validation, and 15\% for testing.  In the summary-level setting, (different) summaries of the same abstract may appear in both the train and the test set.

\xhdr{Post-hoc validation} 
To confirm the validity of our results, we use a set of heuristics to assess whether specific subsets of our data were synthetically generated or human\hyp generated with high precision, relying on the logged keystrokes.
First, we assume that summaries written entirely in the text box made available on MTurk (without involving a pasting action) are real, allowing us to assess whether our above-described classification model (which does not take keystrokes into account) has a low false\hyp positive rate.
Second, for the summaries where pasting was used, we examine which fraction of the pasted text came from the original abstract (as crowd workers simply re-arrange parts of the abstract in their summary), versus which fraction is made up of new text. The fraction of overlap with the original abstract was operationalized as the longest common substring present in both the original abstract and the summary (as in  \citet{suzgun2023string2string}).
Under the assumption that pasted summaries that have very little to do with the original abstract are synthetically generated, we can get a sense of the false negative rate of our model.

\section{Results}
\xhdr{Performance of synthetic-text detector}
We report results for both the summary-level and abstract-level data splits (see Section~\ref{sec:det}) in Table~\ref{tab:finetune_results}. In the summary-level setting, where summaries derived from all abstracts were pooled before splitting the data, our fine-tuned model achieved an accuracy of 99\% and a macro-F1 of 99\%. The high performance indicates a lack of diversity in ChatGPT's summarizations, a known issue in RLHF-trained LLMs~\cite{openai2023gpt4}, and our smaller\hyp capacity detection model seems to be able to pick up on fingerprint\hyp like artifacts introduced by ChatGPT.

The abstract-level split indicates that these artifacts are universal across abstracts. 
Even not seeing a subset of abstracts during training, our model was still capable of successfully identifying synthetic summaries 97\% of the time, with a macro-F1 of 97\%. 
In other words, a ChatGPT\hyp detection model trained on our abstract summarization task only is able to identify new synthetic abstracts almost all of the time.
These high scores indicate that---at least for the task at hand---there exists an identifiable ChatGPT fingerprint in abstract summarization tasks that make it learn universal features to discriminate between real and synthetic texts.

\begin{table*}[t!]
    \centering
    \begin{tabular}{lcccc}
        \toprule
         \textbf{Setting} & \textbf{Accuracy} & \textbf{Macro-F1} & \textbf{Precision} & \textbf{Recall} \\
         \midrule
         Summary-level & 0.99 \tiny{$\pm$0.02} & 0.99 \tiny{$\pm$0.03} & 0.99 \tiny{$\pm$0.02} & 0.98 \tiny{$\pm$0.04} \\
         \midrule
         Abstract-level & 0.97 \tiny{$\pm$0.03} & 0.97 \tiny{$\pm$0.02}& 0.97 \tiny{$\pm$0.02}& 0.97 \tiny{$\pm$0.04}\\
         \bottomrule
    \end{tabular}
    \caption{Test performance of the synthetic-vs.-real classifier in the summary-level and abstract-level setups.}
    \label{tab:finetune_results}
\end{table*}




\begin{figure}
    \centering
    \includegraphics[width=2.5in]{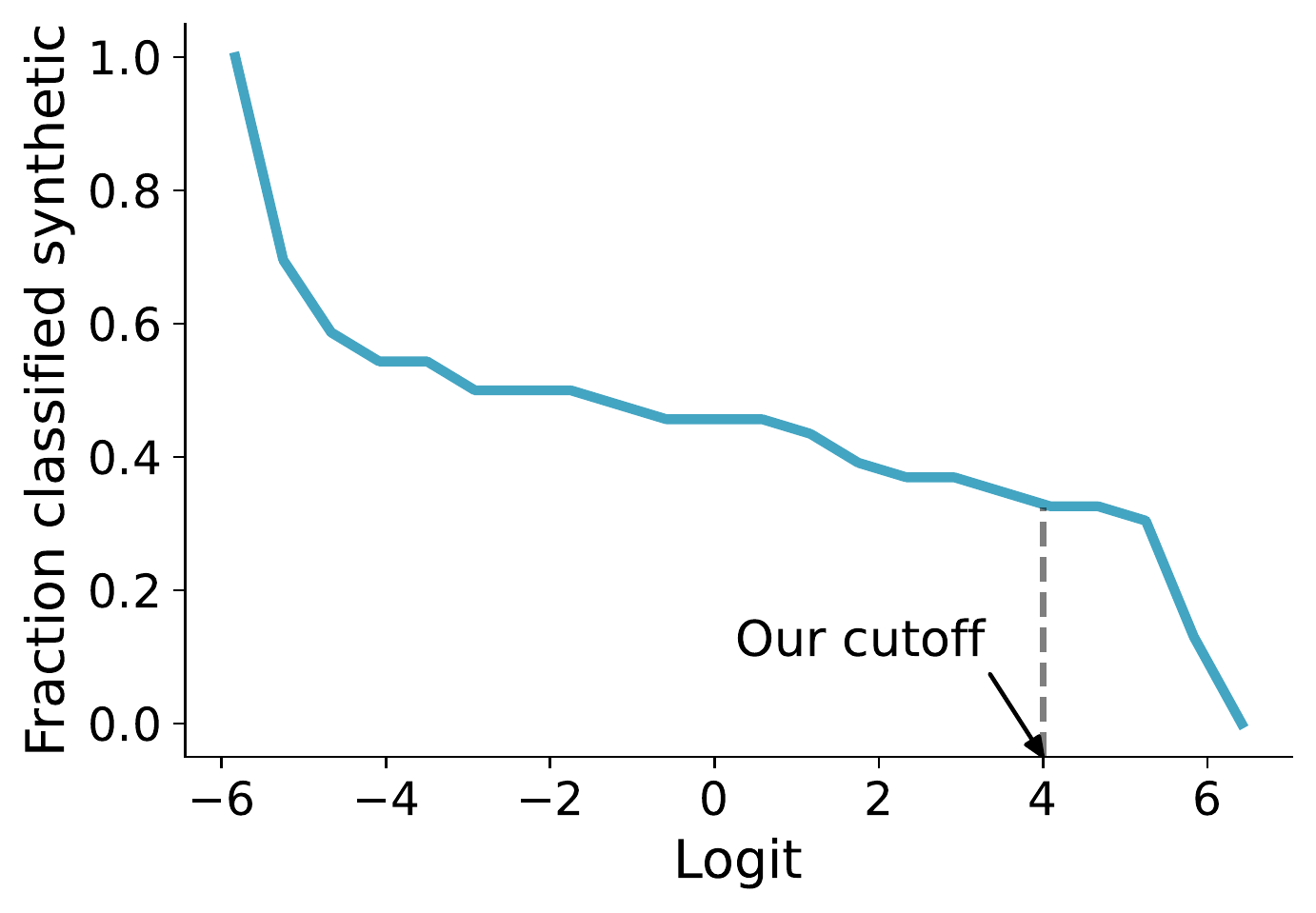}
    \caption{Proportion of summaries predicted as synthetic depending on the logit threshold.}
    \label{fig:logit}
\end{figure}

\xhdr{Prevalence of LLM usage among crowd workers}
Due to the overall higher accuracy, we applied the model trained in the summary-level setting to the 46 new summaries to detect instances of LLM usage among crowd workers.
When running this classifier, we need to choose a decision threshold above which we classify a text as synthetic, and below which we classify it as human\hyp generated.
For robustness, we ran the classifier for a wide range of thresholds and present the fraction of summaries classified as synthetic as a function of the applied threshold in Figure~\ref{fig:logit}.
With a logit threshold of 0 (corresponding to a predicted probability of 50\%), we estimate that 21 of the 46 crowdsourced summaries (\upperestimate{}; 95\% CI: [31\%, 61\%]) were synthetically generated.
In order to avoid misclassifying ambiguous examples as synthetic, we may want to consider a more conservative estimate of the prevalence of LLM usage by choosing a higher logit threshold.
For example, using a threshold of 4 (corresponding to a predicted probability of 98\%), we still arrive at an estimate of 15 of the 46 crowdsourced summaries (33\%; 95\% CI: [20\%, 45\%]) being LLM-generated.
Overall, our estimate remains stable over a wide range of logit thresholds, and we conclude that 33--\upperestimate{} of crowdsourced summaries were produced with the help of LLMs.

Note that, since nearly all users submitted only a single summary (44 workers contributed 46 summaries), the above\hyp mentioned fraction of LLM-produced summaries can also be interpreted as a fraction of LLM-using crowd workers. (Macro-averaging over workers yields the same estimates as the above\hyp reported micro-averages over summaries.)

\xhdr{Post-hoc validation}
Having estimated the prevalence of LLM usage, we conducted a series of analyses to confirm the validity of our approach.
(In these analysis, we use the more conservative logit threshold of 4; see above.)
In Table~\ref{tab:confusion_matrix}, we show that the majority of users pasted at least some text when writing their summaries (affecting 89\%, or 41 out of 46, summaries).
Indeed, only five summaries were written entirely in the text box without any pasting.
Importantly, our classifier labeled all of them as human-written, suggesting the classifier has a low false\hyp positive rate (under the aforementioned assumption that these summaries were human-generated).

\begin{figure}[t!]
    \centering
    \includegraphics[width=2.6in]{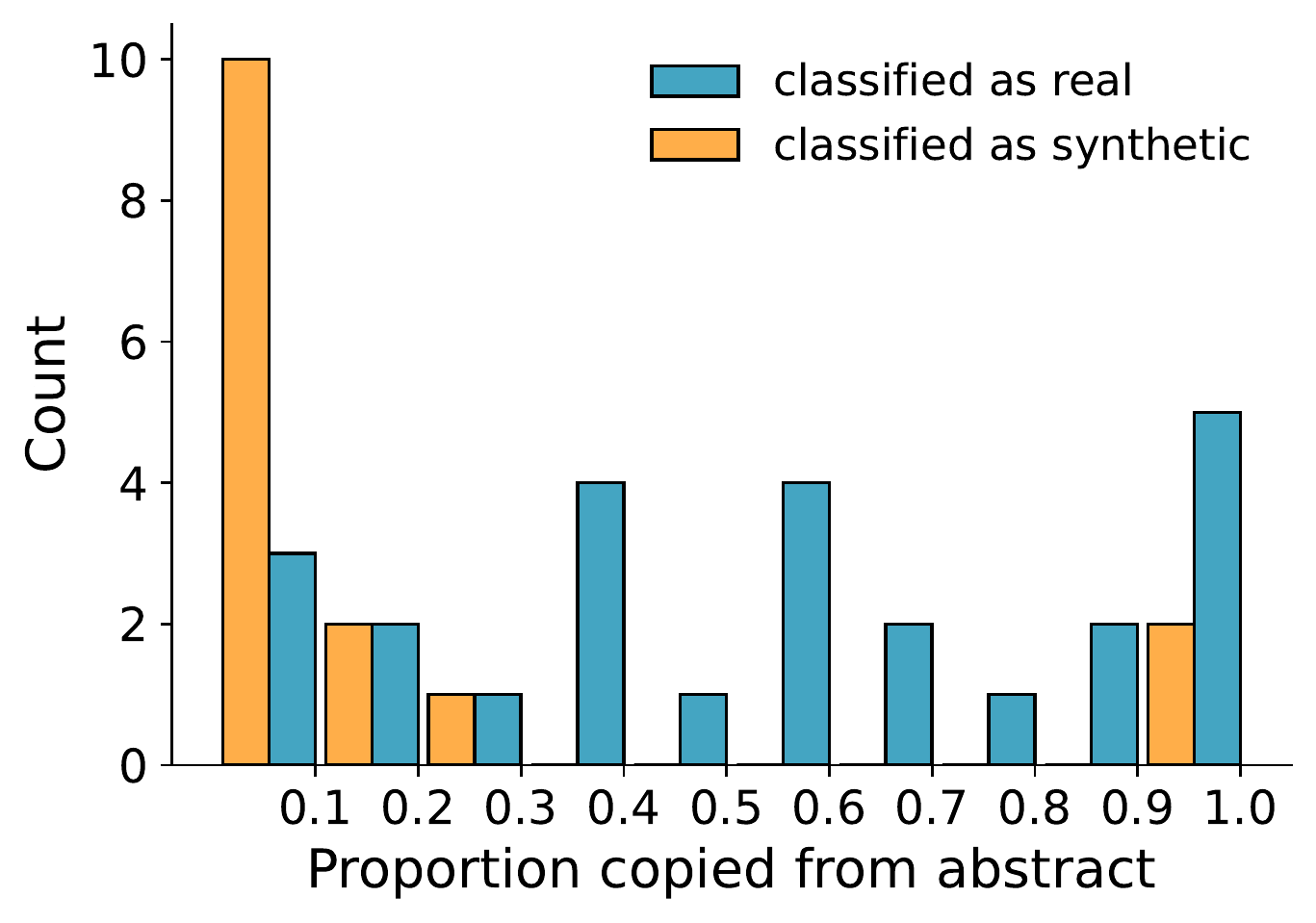}
    \caption{%
    Overlap between summaries and original abstracts (operationalized as ratio of length of longest common substring and length of original abstract), for summaries involving a paste action.
    }
    \label{fig:proportion_copied}
\end{figure}

The mere act of pasting does not imply the usage of ChatGPT. 
In particular, a qualitative analysis of the data at hand suggests that workers often copy--paste intricate phrasing, or entire abstracts, from the original content into their text editor, thereby reutilizing abstract content.
Thus, to understand how copy--pasting was used in summaries classified as synthetic vs.\ human, we compute the longest common substring between each summary and the original abstract (henceforth ``overlap''). 
Our analysis, depicted in Figure~\ref{fig:proportion_copied}, reveals that workers often reuse large portions from the original abstracts, but also that, more importantly, summaries classified as synthetic mostly had a small overlap with the original abstracts. For instance,
out of 13 summaries with an overlap of less than 10\%, 10 (76\%) were classified as synthetically generated.
Also, most summaries classified as synthetic have a small overlap with the original abstract.
This suggests that what was being copy--pasted was not parts from the original abstract but, indeed, outputs from an LLM.

\begin{table}[t!]
    \centering
    \begin{tabular}{r|cc}
         & \textbf{With pasting} & \textbf{Without pasting} \\
         \midrule
         \textbf{Synthetic} & 15 & 0 \\
         \textbf{Human} & 26 & 5
    \end{tabular}
    \caption{Matrix showing the link between usage of pasting (columns) and classifier decision (rows). Cells contain the number of summaries labeled as synthetic\slash human-written in whose production pasting was\slash was not used.}
    \label{tab:confusion_matrix}
\end{table}

\section{Discussion}
The pivotal role of human-generated data in various applications is undeniable. Its richness, uniqueness, and diversity are crucial factors that make it stand apart from synthetically generated data~\cite{veselovsky2023generating, ziems2023can}.
Here we found that crowd workers on MTurk widely use LLMs in a summarization task,  which raises serious concerns about the gradual dilution of the ``human factor'' in crowdsourced text data.
Further, we developed a robust, low-computational-cost method for synthetic\hyp data detection, indicating that bespoke detection models may be more useful than out-of-the-box solutions.
Our setup maps well to other settings where synthetically generated text may be problematic, including the education space, where tests, essays, and assignments can be quickly and often effectively solved by LLMs~\cite{rudolph2023war}.
 

There is widespread concern that LLMs will shape our information ecosystem, i.e., that much of the information available online will be created by LLMs. This may degrade the performance of ``recursive'' LLMs, i.e., those trained on synthetically generated data~\cite{shumailov2023curse}, and magnify the impact of the values and ideologies encoded by these models~\cite{santurkar2023whose,liu2022wanli}. In that vein, the present work raises the concern that acquiring human data may be made even harder with the popularization of LLMs, as crowd workers seem to already be using it extensively, a problem that could become much worse with the popularization of LLMs and the increase in their capabilities.




All this being said, we do not believe that this will signify the end of crowd work, but it may lead to a radical shift in the value provided by crowd workers. Instead of providing \textit{de novo} annotations, crowd workers may instead serve as an important human filter for detecting when these models succeed and when they fail. 
Early work has already made significant progress in this direction, pairing humans and language models to create high-quality, diverse data~\cite{liu2022wanli}.

\xhdr{Limitations}
In this study, our focus is limited to one specific crowdsourcing task: text summarization. 
While summarization  captures many of the nuances needed for text production tasks in general, we acknowledge the uncertainty regarding the generalization of our findings to other tasks, particularly those that pose substantial challenges to LLMs.
This highlights an important area for future research, which involves comprehensively examining how the results may vary across different tasks and how they evolve over time as LLMs become even more widespread.
Nonetheless, we speculate that the phenomenon uncovered here may affect any text production task that is specified via a textual instruction that can be readily used as a prompt for an LLM, and
our findings should certainly serve as a cautionary tale for researchers and practitioners working with other types of data and tasks.

\section{Ethical considerations}
Our study used keystroke collection to validate the results. Though beneficial for research, keystroke tracking could potentially infringe upon user privacy if not appropriately handled. We strictly limited the keystroke tracking to users' interactions with the edit box and copy-pastes for the page. However, we believe that more expansive use of tracking can be problematic.

\bibliography{00MainPaper}
\balance
\bibliographystyle{acl_natbib}

\end{document}